\documentclass[11pt,a4paper]{article}
\usepackage[hyperref]{emnlp2018}
\usepackage{times}
\usepackage{latexsym}
\usepackage{url}
\usepackage{multicol}
\usepackage{graphicx}
\usepackage{amsmath}
\usepackage{amssymb}
\usepackage{algorithm}
\usepackage{algpseudocode}
\usepackage{multirow}

\newcommand*\Let[2]{\State #1 $\gets$ #2}

\aclfinalcopy % Uncomment this line for the final submission

\title{Extending Event Detection to New Types with Learning from Keywords}

\author{Viet Dac Lai {\normalfont and} Thien Huu Nguyen \\
  Department of Computer and Information Science  \\
  University of Oregon, OR, USA \\
  {\tt \{vietl,thien\}@cs.uoregon.edu} \\
}

\date{}

\begin{document}
\maketitle

\begin{abstract}
Traditional event detection classifies a word or a phrase in a given sentence for a set of predefined event types. The limitation of such predefined set is that it prevents the adaptation of the event detection models to new event types. We study a novel formulation of event detection that describes types via several keywords to match the contexts in documents. This facilitates the operation of the models to new types. We introduce a novel feature-based attention mechanism for convolutional neural networks for event detection in the new formulation. Our extensive experiments demonstrate the benefits of the new formulation for new type extension for event detection as well as the proposed attention mechanism for this problem.

\end{abstract}

% <concept_id>10010147.10010178.10010179.10003352</concept_id>
% <concept_desc>Computing methodologies~Information extraction</concept_desc>
% <concept_id>10010147.10010257.10010293.10010294</concept_id>
% <concept_desc>Computing methodologies~Neural networks</concept_desc>

% \ccsdesc[500]{Computing methodologies~Information extraction}
% \ccsdesc[500]{Computing methodologies~Neural networks}

%
% Keywords. The author(s) should pick words that accurately describe the work being
% presented. Separate the keywords with commas.
% \keywords{}
%
% This command processes the author and affiliation and title information and builds
% the first part of the formatted document.
\maketitle

\section{Introduction}

Event detection (ED) is a task of information extraction that aims to recognize event instances (event mentions) in text and classify them into specific types of interest. Event mentions are usually associated with an event trigger/anchor in the sentence of the event mentions, functioning as the main word to evoke the event. For instance, in the sentence \textit{"She is going to leave to become chairman of Time Inc."}, an ED system should be able to recognize that the word ``{\it leave}'' is triggering an event of type ``{\it End-Position}''.

There have been two major approaches for ED in the literature. The first approach focuses on the development of linguistic features to feed into the statistical models (i.e., MaxEnt) \cite{Ahn:06,Ji:08,Liao:10,McClosky:11}. The second approach, on the other hand, relies on deep learning (i.e., convolutional neural networks (CNN)) to automatically induce features from data \cite{Chen:15,Nguyen:16,Liu:17,Lu:18}, thus significantly improving the performance for ED.

One limitation of the current approaches for ED is the assumption of a predefined set of event types for which data is manually annotated to train the models. For example, the popular benchmark dataset ACE 2005 for ED annotates 8 types and 33 subtypes of events. 
Once the models have been trained in this way, they are unable to extract instances of new, yet related types (i.e., having zero performance on the new types). To extend the operation of these models into the new types, the common approach is to spend some effort annotating data for the new types to retrain the models. Unfortunately, this is an expensive process as we might need to obtain a large amount of labeled data to adequately represent various new event types in practice. Such expensive annotation has hindered the application of ED systems on new types and calls for a better way to formulate the ED problem to facilitate the extension of the models to new event types. 

In this paper, we investigate a novel formulation of ED where the event types are defined via several keywords instead of a large number of examples for event types in the traditional approaches (called the learning-from-keyword formulation (LFK)). These keywords involve the words that can possibly trigger the event types in the contexts. For instance, the event type {\it End-Position} can be specified by the keywords (``{\it left}, ``{\it fired}'', ``{\it resigned}''). Given the keywords to represent event types, the ED problem becomes a binary classification problem whose goal is to predict whether a word in a sentence expresses the event type specified by the keywords or not. This formulation enables the ED models to work with new event types as long as the keywords to describe the new types are provided, thus allowing the ED models to be applicable on a wide range of new event types and mitigating the needs for large amounts of annotated data for the new types.

The goal of this paper is to evaluate the effectiveness of LFK in the new type extension setting for ED where the models are trained on labeled data from some types but applied to extract instances of unseen types. We would like to promote this problem as a new task for ED for future research. To set the baselines for this problem, we employ the ACE 2005 dataset and recast it into LFK. We examine the performance of the baseline models for ED in the traditional formulation when they are adapted to LFK. The experiments show that with the new formulation, such ED models can actually recognize new event types although their performance should be still further improved in future research. Finally, we demonstrate one possibility to improve the performance of the baseline ED models in LFK by presenting a novel attention mechanism for CNNs based on the feature space to fuse the representations of the keywords and contexts. We achieve the state-of-the-art performance for the new type extension with the proposed attention mechanism.

\section{Related work}
In the last decade, many machine learning systems have been introduced to solve ED. Before the era of the deep neural networks, these systems are mainly based on supervised learning using extensive feature engineering with the machine learning frameworks \cite{Ahn:06,Ji:08,Hong:11,Riedel:09,Riedel:11a,Riedel:11b,Makoto:14,Li:14,Li:15}. Recently, many advanced deep learning methods were introduced to enhance event detectors such as distributed word embedding \cite{Chen:15,Nguyen:16a:joint,Liu:17,nguyen:18:one}, convolutional neural networks \cite{Chen:15,Chen:17,Nguyen:15:event,Nguyen:16a:joint,nguyen:18:graph}, recurrent neural networks \cite{Nguyen:16a:joint,Sha:18}, and the attention mechanism \cite{Liu:17,Nguyen:18b,Liu:18}. However, the models proposed in these work cannot extend their operation to new event types.

Regarding the new formulations for ED, previous studies\cite{bronstein:15:seed,Peng:16:minimal} also examine keywords to specify event types. However, these studies do not investigate the new type extension setting as we do in this. Recently, zero-shot learning is employed for new types in event extraction\cite{huang:18:zeroshot}; however, the event types are specified via the possible roles of the arguments participating into the events in this work. It also uses complicated natural language processing toolkits, making it difficult to apply and replicate the settings. Our work emphasizes the simplicity in the setting for new type extension to facilitate future research. Finally, extending ED to the new type is investigated using real examples as new event types \cite{Nguyen:16c:twostage}. However, it requires a large number of examples to perform well. Our work instead requires only a few keywords to help the models achieve reasonable performance on new types.

\section{Learning-from-Keywords for ED}
\subsection{Task Definition}

In the learning-from-keyword formulation for ED, the inputs include a context (i.e., an $n$-word sentence $X=\{x_1, x_2, \ldots, x_n\}$ with an anchor word located at position $a$ (the word $x_a$)) and a set of keywords $K$. The words in $K$ are the possible trigger words of some event type of interest. The goal is to predict whether the word $x_a$ in $S$ expresses the event type specified by $K$ or not (i.e., a binary classification problem to decide whether the context matches the event keywords or not). An example in LFK thus has the form $(X, x_a, K, Y)$ where $Y$ is either 1 or 0 to indicate the match of $X$ and $K$.

\subsection{Data Generation}
\label{sec:data-gen}

To facilitate the evaluation of the ED models in LFK for the new type extension setting, we need to obtain training and test/development datasets so the keyword sets of the examples in the test/development datasets define event types that are different from those specified by the keyword sets in the training datasets. To our best knowledge, there is no existing data following LFK setting, therefore, in this section, we present a process to automatically generate an ED dataset for LFK setting from an existing ED dataset.

We obtain these datasets by leveraging ACE 2005, the popular benchmark datasets for ED. ACE 2005 dataset is annotated for 8 event types $\mathcal{T} = \{t_1, t_2,\ldots, t_8\}$, and 33 event subtypes $\mathcal{S} = \{s_1, s_2, \ldots, s_{33}\}$. There is also a special type/subtype of ``{\it Other}'' indicating the non-event instances ($\textit{Other} \notin \mathcal{S}$) . As each event subtype in ACE 2005 is associated with one event type, let $\mathcal{C}_i$ be the set of subtypes corresponding to the type $t_i \in \mathcal{T}$. Also, let $\mathcal{K}_j$ be the set of trigger words for the event mentions of the subtype $s_j \in \mathcal{S}$. $\mathcal{K}$ is collected from training set of ACE 2005.

%that do not express any type in $\mathcal{T}$ and $\mathcal{S}$

To generate the training and test/development datasets, we first split the documents in ACE 2005 into three parts $D_{train}$, $D_{test}$ and $D_{dev}$ following the previous work on ED \cite{Li:13}. They would contain event mentions for all the possible event types and subtypes in $\mathcal{T}$ and $\mathcal{S}$. Assume that we want to extend the system to a new event type $t_{target} \in \mathcal{T}$, we need a train set without $t_{target}$. So, we remove every event mention whose subtype belongs to $\mathcal{C}_{target}$ from $D_{train}$. Whereas, samples with subtypes in $\mathcal{C}_{target}\cup\{Other\}$ are kept in $D_{test}$ and $D_{dev}$. The results of this removal process are called as $D'_{train}$, $D'_{test}$ and $D'_{dev}$ (from $D_{train}$, $D_{test}$ and $D_{dev}$, respectively). They will be used to generate the actual training/test/development datasets for LFK, respectively.

Specifically, for each of these datasets (i.e., $D'_{train}$, $D'_{test}$ and $D'_{dev}$), the goal is to produce the positive and negative examples in corresponding LFK datasets. Algorithm \ref{alg} shows the pseudo-code to generate the training dataset for LFK from $D'_{train}$. The same algorithm can be applied for the test and development dastasets of LFK, but replace $D'_{train}$ with $D'_{test}$ and $D'_{dev}$ respectively in line 2, and replace $\mathcal{S}\setminus \mathcal{C}_{target}$ with $\mathcal{C}_{target}$ in line 10. 

\let\oldReturn\Return
\renewcommand{\Return}{\State\oldReturn}

\begin{algorithm}
\caption{Training dataset generation for LFK}
\label{alg}
\begin{algorithmic}[1]

\Let{$D^{+}_{train}, D^{-}_{train}$}{$\emptyset,\emptyset$} \Comment{Positive and negative example sets}

\For{$(X, x_a, s_j) \in D'_{train}$}
\Comment{where $X$ : a sentence, $x_a \in X$ : the anchor word, $s_j \in \mathcal{S}$ : the corresponding subtype}

\If{$s_j \ne$ ``{\it Other}''}
% \Let{$s$}{$s_i \in \mathcal{S}$}
\For{$u = 1..5$}
\Let{$K^u_j$}{A subset of $\mathcal{K}_j \setminus \{x_a\}$: $|K^u_j| = 4$}
\Let{$D^{+}_{train}$}{$D^{+}_{train} \cup \{(X, x_a, K^u_j, 1)\}$}
\EndFor

%Sample 5 set of words $K_j$
%so $|K_j| = 4$, $K_j \subset \mathcal{K}_i$

\Else \Comment{$s =$ ``{\it Other}''}
\Let{$s_v$}{Some subtype in $\mathcal{S} \setminus \mathcal{C}_{target}$}
\Let{$K$}{A subset of $\mathcal{K}_v$: $|K| = 4$}
\Let{$D^{-}_{train}$}{$D^{-}_{train} \cup \{(X, x_a, K, 0)\}$}
\EndIf

\EndFor
\Return{$D^{+}_{train}$ and $D^{-}_{train}$}
\end{algorithmic}

\end{algorithm}

Since the number of positive examples in $D_{test}$ set is small, we choose two event types (i.e., {\it Conflict} and {\it Life}) that have the largest numbers of positive examples in $D_{test}$ as the target types. Applying the data generation procedure above, we generate a dataset in LFK for each of these target types. 

\section{Model}

This section first presents the typical deep learning models in the traditional ED formulation adapted to LFK. We then introduce a novel attention mechanism to improve such models for LFK. 

\subsection{Baselines}

As CNNs have been applied to the traditional formulation of ED since the early day \cite{Chen:15,Nguyen:15:event,Nguyen:16b:modeling}, we focus on the CNN-based model in this work and leave the other models for future research.

\textbf{Encoding Layer}: To prepare the sentence $S$ and the anchor $x_a$ for the models, we first convert each word $x_i \in S$ into a concatenated vector $h^0_i = [p_i,q_i]$, in which $p_i \in \mathbb{R}^u$ is the position embedding vector and $q_i \in \mathbb{R}^d$ is the word embedding of $x_i$. We follow the settings for $p_i$ and $q_i$ described in \cite{Nguyen:15:event}. This step transforms $S$ into a sequence of vector $H^0 = (h^0_1, h^0_2, \ldots, h^0_n)$.

\textbf{Convolution Layers}: Following \cite{Chen:15,Nguyen:15:event}, we apply a convolutional layers with multiple window sizes for the filters $W$ over $H_0$, resulting in a sequence of hidden vectors $H^1 = (h^1_1, h^1_2, \ldots, h^1_n)$. Note that we pad $H^0$ with zero vectors to ensure that $H^1$ still has $n$ vectors. We can essentially run $m$ convolutional layers in this way that would lead to $m$ sequences of hidden vectors $H^1, H^2, \ldots, H^m$.

\textbf{Keyword Representation}: We generate the representation vector $V_K$ for the keyword set $K$ by taking the average of the embeddings of its words.

Given the keyword vectors $V_K$ and the hidden vector sequences from CNNs for $S$ (i.e., $H^1, H^2, \ldots, H^m$), the goal is to produce the final representation $R = G(V_K, H^1, H^2, \ldots, H^m)$, serving as the features to predict the matching between $(S, x_a)$ and $K$ (i.e., $R$ would be fed into a feed-forward neural network with a softmax layer in the end to perform classification). There are two immediate baselines to obtain $R$ adapted from the models for traditional ED:

(i) {\it Concat}: In this method, we apply the usual max-pooling operation over the hidden vectors for the last CNN layer $H^m$ whose result is concatenated with $V_K$ to produce $R$ \cite{Nguyen:15:event,Chen:15}.

(ii) {\it Attention}: This method applies the popular attention method to aggregate the hidden vectors in $H^m$ using $V_K$ as the query \cite{Bahdanau:15}. The formulas for $R$ are shown below:
\begin{equation*}
\begin{split}
u_i & = \sigma(W_u h^m_i + b_u)\\
c &= \sigma(W_c[V_K, h^m_a] + b_c) \\
\alpha_i & = \frac{\exp(c^{\top} u_i)}{\sum_{j}\exp(c^{\top} u_j)}\\
R &= \sum_{i} \alpha_i h^m_i
\end{split}
\end{equation*}

\subsection{Conditional Feature-wise Attention}

The interaction between the keywords and hidden vectors in the baselines is only done in the last layer, letting the intermediate CNN layers to decide the computation themselves without considering the information from the keywords. 

To overcome this limitation, we propose to inject supervision signals for each CNN layer in the modeling process. In particular, given the sequence of hidden vectors $H^i = (h^i_1,h^i_2,\ldots,h^i_n)$ obtained by the $i$-th CNN layer, instead of directly sending $H^i$ to the next layer, we use $V_K$ to generate the representation vectors $\gamma_i$ and $\beta_i$, aiming to reveal the underlying information/constraints from the keywords that the $i$-th CNN layer should reason about. Such representation vectors condition and bias the hidden vectors in $H^i$ toward the keywords based on the feature-wise affine transformation \cite{perez:17:film}. The conditioned hidden vectors from this process (called $\bar{H}^i = (\bar{h}^i_1,\bar{h}^i_2,\ldots,\bar{h}^i_n)$) would be sent to the next CNN layer where the conditional process guided by the keywords continues:
\begin{equation*}
\begin{aligned}
\gamma_i &= \sigma(W^i_{\gamma}V_K + b^i_{\gamma}) \\
\beta_i &= \sigma(W^i_{\beta}V_K + b^i_{\beta}) \\
\bar{h}^i_j &= \gamma_i * h^i_j + \beta_i
\end{aligned}
\end{equation*}
where $\sigma$ is a non-linear function while $W^i_{\gamma}, b^i_{\gamma}, W^i_{\beta}$ and $b^i_{\beta}$ are model parameters.

We call the operation described in this section the Conditional Feature-wise Attention (CFA). The application of CFA into the two baselines \textbf{Concat} and \textbf{Attention} leads to two new methods \textbf{Concat-CFA} and \textbf{Attention-CFA} respectively.

\section{Experiments}
We use the datasets generated in Section \ref{sec:data-gen} to evaluate the models in this section. Table \ref{tbl:data} shows the staticstics of our generated datasets. This dataset will be publicly available to the community.

\begin{table}[h!]
\centering
    \begin{tabular}{|l| c | r r r |}
         \hline
         & Label & Train & Dev & Test \\
         \hline
         \multirow{2}{*}{Conflict} & +1 &  14,749 & 929 & 509 \\
         & -1 & 177,421 & 13,130 & 13,576 \\
         \hline
         \multirow{2}{*}{Life} & +1 & 17,434 & 354 & 154 \\
         & -1 & 177,421 & 13,130 & 13,576 \\
         \hline
    \end{tabular}
    \caption{Numbers of the positive and negative samples of the LFK datasets.}
    \label{tbl:data}
\end{table}

\subsection{Parameters}

\begin{table*}[h!]
\centering
\begin{tabular}{|l|c c c|c c c|}
    \hline
     \multirow{2}{*}{Model} & \multicolumn{3}{c|}{Conflict} & \multicolumn{3}{c|}{Life} \\
     %\multicolumn{3}{c}{Movement}  \\
     & P & R & F1 & P & R & F1 %
     %& P & R & F1\\
     \\
     \hline
     Feature & 21.7 & 9.8 & 13.5 &14.4 & 25.8 & 18.5
     \\
     Word2Vec & 20.6 & 72.4 & 32.1 & 4.4 & 61.9 & 8.2
     \\
     Feature + Word2vec & 27.8 & 20.2 & 23.4 & 15.7 & 31.6 & 21.0
     \\
     Seed & 11.9 & 36.1 & 17.9 & 9.5 & 71.0 & 16.7
     \\
     \hline
    %  Concat &  20.5 & 57.8 & 30.0 & 10.9 & 48.3 & 17.7
     Concat &  20.5 & 57.8 & 30.0 (4) & 10.9 & 48.3 & 17.7 (2)
     %& 1477 & 6392 & 2362\\
     \\
     Attention &  21.5 & 59.1& 31.4 (4)& 12.8 & 45.0 & 19.1 (2)
    %  Attention &  21.5 & 59.1& 31.4& 12.8 & 45.0 & 19.1
     %& 1439 & 6450 & 2332\\
     \\
    %  Concat-CFA &  25.1 & 57.1 & 33.8 & 10.6 & 43.6 & 16.9
     Concat-CFA &  25.1 & 57.1 & 33.8 (4)& 10.6 & 43.6 & 16.9 (1)
     %& 1434 & 3967 & 2044\\
     \\
    %  Attention-CFA &  22.5 & 74.2 & \textbf{34.1} &  18.5 & 38.7 & \textbf{25.0}
     Attention-CFA &  22.5 & 74.2 & \textbf{34.1} (1) &  18.5 & 38.7 & \textbf{25.0} (4)
       %1131 & 7300 & 1932\\
       \\
     \hline
\end{tabular}
\caption{Model performance. The numbers in the brackets indicate the optimized numbers of CNN layers.}
\label{tbl:test}
\end{table*}

We examine four deep learning models: baselines (i.e., {\it Concat} and {\it Attention}) and the proposed models (i.e., {\it Concat-CFA} and {\it Attention-CFA}). Following \cite{Nguyen:15:event}, we employ the {\tt word2vec} word embeddings from \cite{Mikolov:13} with 300 dimensions for the models in this work. The other parameters for the deep learning models in this work are tuned on the development datasets. In particular, we employ multiple window sizes (i.e., 2, 3, 4 and 5) in the CNN layers, each has 100 filters. We use Adadelta as the optimizer with the learning rate set to 1.0. We apply a dropout with a rate of 0.5 to the final representation vector $R$. Finally, we optimize the number of CNN layers for each deep learning model. %The actual value for each model is shown in Table \ref{tbl:test}.

In addition, we investigate the typical feature-based models with the MaxEnt classifier in the traditional ED formulation for LFK to constitute the baselines for future research. In particular, we examine four feature-based models for ED in LFK:
\begin{itemize}
    \item \textbf{Feature} combines the state-of-the-art feature set for ED designed in \cite{Li:13} for the input context ($S, x_a$) with the words in the keyword set $K$ to form its features
    \item \textbf{Word2Vec} utilizes the keyword representation $V_K$ and the average of the embedding of the words in the window size of 5 for $x_a$ in $S$ as the features
    \item \textbf{Feature + word2vec} uses the aggregated features from above models
    \item \textbf{Seed} employs the model with semantic features in \cite{bronstein:15:seed}.
\end{itemize}

\subsection{Evaluation}

Table \ref{tbl:test} presents the performance of the models on the test performance for different datasets (i.e., with {\it Conflict} and {\it Life} as the target type). Among the features in the feature-based models, the embedding features in {\it Word2Vec} are very helpful for ED in LFK as the models with these features achieve the best performance (i.e., {\it Word2Vec} for {\it Conflict} and {\it Feature+Word2Vec} for {\it Life}). Among the deep learning models, the CFA-based models (i.e., {\it Concat-CFA} and {\it Attention-CFA}) are significantly better than their corresponding baseline models (i.e., {\it Concat} and {\it Attention}) over both {\it Conflict} and {\it Life} with {\it Attention}. This confirms the benefits of CFA for ED in LFK.

Comparing the deep learning and the feature-based models, it is interesting that the feature-based models with average word embedding features can perform better than the deep learning baseline models (i.e., {\it Concat} and {\it Attention}) for {\it Conflict}. However, when the deep learning models are integrated with both attention and CFA (i.e., {\it Attention-CFA}), it achieves the best performance over both datasets. This helps to testify to the advantage of deep learning and CFA for ED in the new type extension setting with LFK.

Finally, although the models can extract event mentions of the new types, the performance is still limited in general, illustrating the challenge of ED in this setting and leaving many rooms for future research (especially with deep learning) to improve the performance. We hope that the setting in this work presents a new way to evaluate the effectiveness of the ED models.

\section{Conclusion}

We investigate a new formulation for event detection task that enables the operation of the models to new event types, featuring the use of keywords to specify the event types on the fly for the models. A novel feature-wise attention technique is presented for the CNN models for ED in this formulation. Several models are evaluated to serve as the baselines for future research on this problem.

\bibliographystyle{acl_natbib_nourl}
\bibliography{emnlp2018.bbl}
\end{document}